\documentclass{llncs}
\usepackage{llncsdoc}
\usepackage[dvips]{graphicx}
\begin{document}

\title{Towards an Error Correction Memory to Enhance Technical Texts Authoring in LELIE}
\author{Juyeon Kang (2), Patrick Saint-Dizier (1)}
\institute{ (1) IRIT - CNRS, (2) Prometil\\
Toulouse, France.\\
j.kang@prometil.com, stdizier@irit.fr}

\maketitle

{\bf Abstract}

In this paper, we investigate and experiment the notion of error correction memory applied to error correction in technical texts. The main purpose is
to induce relatively generic correction patterns associated with more contextual correction recommendations, based on previously memorized and analyzed corrections.
The notion of error correction memory is developed within the framework of the LELIE project and
illustrated on the case of fuzzy lexical items, which is a major problem in technical texts.

\section{Introduction}

Technical documents form a linguistic genre with specific linguistic constraints in terms of lexical realizations, including business or domain dependent aspects, 
grammar and style. 
These documents are designed to be easy to read and as efficient and unambiguous as possible for their users and readers. 
For that purpose, they tend to follow relatively strict controlled natural language principles concerning both their form and contents. Guidelines
for writing in controlled languages have been elaborated in various sectors, they are summarized in e.g. (Alred 2012), (Umwalla 2004), (O'Brian 2003), (Weiss 2000), and (Wyner et al. 2010). 
Besides guidelines, the boilerplate technique is also used for simple texts or for requirement authoring.

Authoring principles and guidelines, in the everyday life of technical writers, are often only partly observed, for several reasons including workload and 
the large number of revisions
made by several actors on a text. Table 1 below shows some major errors found by the LELIE system over 300 pages of technical documentation for companies A, B and C (kept anonymous) in spite of
the strict guidelines they impose. These results show that there are still many errors of various types and space for correction strategies.

In the LELIE project (Barcellini et al. 2012), we developed a system that detects several types of errors in technical documents and produces {\bf alerts}.
The LELIE system makes local parses of technical texts in order to detect writing errors. Parses ranges from finding fuzzy terms to complex structures that must be
revised (e.g. discourse structures, coordinations of NPs). In both cases, it is necessary to develop some kind of 'local' grammar to recognize ill-formed constructions,
but also to filter out others which are correct (e.g. fuzzy terms in business terms are correct).

The alerts produced by the LELIE system have been found useful by most technical writers that tested the system. However, to be really helpful to technical writers,
it turns out that (1) false positives (about 30\% of the alerts) must be filtered out and (2) help must be provided to technical writers under the form of 
correction patterns and recommendations whenever possible.
Our ongoing research aims at specifying, developing and testing several facets of an {\bf error correction memory} system that would, after
a period of observation of technical writers making corrections from the LELIE alerts, (1) memorize errors which are not
or almost never corrected so that they are no longer displayed in texts in the future and (2) memorize corrections realized by writers and propose 
typical correction recommendations.
Our approach is aimed at being very flexible w.r.t. the writer's practices. It is more flexible than systems such as RAT-RQA, Rubric, Attempto, Peng, or Rabbit which
are based the recognition of fixed erroneous structures.

{\small
\begin{center}
\begin{tabular}{|l|l|c|c|c|}
\hline
error type & frequency / 1000 lines  & A & B & C \\  \hline \hline
fuzzy lexical items & 66 & 44 & 89 & 49  \\ \hline 
modals in instructions & 5 & 0 & 12 & 1 \\ \hline 
pronouns with unclear reference & 22 & 4 & 48 & 2  \\ \hline 
negation & 52 & 8 & 109 & 9  \\ \hline 
complex discourse structures & 43 & 12 & 65 & 50  \\ \hline
complex coordinations & 19 & 30 & 10 & 17  \\ \hline 
heavy N+N or noun complements & 46 & 58 & 62 & 15  \\ \hline 
passives & 34 & 16 & 72 & 4  \\ \hline 
sentences too complex & 108 & 16 & 221 & 24  \\ \hline 
incorrect references & 13 & 33 & 22 & 2  \\ \hline 
\end{tabular}

Table 1. Errors found in technical texts for companies A, B and C
\end{center}}

In this paper, we develop elements of a method that shows how to construct (1) relatively generic {\bf correction patterns} paired with (2)  accurate {\bf contextual correction 
recommendations}, based on previously memorized and analyzed corrections. 
The approach of a correction memory that helps technical writers by providing them with error corrections validated and made homogeneous over a whole team of technical
writers, via discussion and mediation, seems to be new to the best of our knowledge. 

This notion of error correction memory originates from the notion of translation memory, it is however substantially different in its
principles and implementation. An in-depth analysis of memory-based language processing is developed (Daelemans et al. 2005) and implemented in the TiMBL software.
This work develops several forms of statistical means to produce generalizations in syntax, semantics and morphology. It also warns against excessive forms of generalizations.
(Buchholz, 2002) develops an insightful memory-based analysis on how grammatical constructions can be induced from samples. Memory-based systems are also used to resolve ambiguities,
using notions such as analogies (Schriever et al. 1989). Finally, memory-based techniques are used in programming languages support systems to help programmers to resolve frequent errors.

\section{The case of fuzzy lexical items}

The LELIE system (features and performances are given in (Barcellini et al. 2012), (Saint-Dizier 2014)) detects several types of errors, lexical, syntactic and related
to style. It also allows to specify business constraints such as controls on style and the use of business terms. The errors detected by LELIE are typical errors of
technical texts (e.g. Table 1), they would not be errors in ordinary language. 
Error detection in LELIE depends on the discourse structure: for example modals are the norm in requirements but not in instructions. Titles allow deverbals which are
not frequently admitted in instructions or warnings. The output of LELIE is the original text with annotations. 
LELIE is parameterized and offers several levels of alerts depending on the a priori error severity. LELIE and the experiments reported below are developed on the
logic-based $<$TextCoop$>$
platform (Saint-Dizier 2012). Lelie is fully implemented and is freely available.

Let us now focus in this short article on the case of fuzzy lexical items which is a major type of error, quite representative of what an error correction memory could be.
Roughly, a fuzzy lexical item denotes a concept whose meaning, interpretation, or boundaries can vary considerably according to context, readers 
or conditions, instead of being fixed once and for all.
It is important to note that (1) that it is difficult
to precisely define and identify what a fuzzy lexical item is (to be contrasted in our context with vague and underspecified expressions, which involve different forms of corrections) 
and (2) that
there are several categories of fuzzy lexical items. These categories include adverbs (manner, temporal, location, and modal adverbs), adjectives ({\it adapted, appropriate}) 
determiners ({\it some, a few}), prepositions ({\it near, around}), a few verbs ({\it minimize, increase}) and nouns. These categories are not homogeneous in terms of 
fuzziness, e.g. determiners and prepositions are always fuzzy. The degree of fuzziness is also quite different from one term to 
another in a category. 
Note that a verb such as {\it damaged} in {\it the mother card risks to be damaged} is not fuzzy but vague because the importance and the nature of the damage is unknown;
{\it heat the probe to reach 500 degrees} is not fuzzy but underspecified because the means to heat the probe are not
given: an adjunct is missing in this instruction. 

The context in which a fuzzy lexical item is uttered may also have an influence on its severity level. For example 'progressively' 
used in a short action ({\it progressively close the water pipe}) or used in an action that has a substantial length ({\it progressively heat the probe till 
300 degrees Celsius are reached}) may entail different severity levels because the application of 'progressively' may be more difficult to realize 
in the second case.  This motivates the need to memorize the context of the error to establish an accurate error diagnosis.

In average, a fuzzy lexical item is found every 4 sentences in our corpus. In our test corpus, from 420 manually annotated fuzzy lexical items, 
LELIE has a detection precision of 88\% with 11\% of noise.
Then, on a smaller experiment with two technical writers from the 'B' company, 
considering 120 different fuzzy lexical items used in different contexts, 36 have been judged not to be errors (30\%): they are noise or minor problems. 
Among the other 84 errors, only 62 have
been corrected. The remaining 22 have been judged problematic and very difficult to correct. It took between 2 and 10 minutes to correct each of the 62 errors, with an average
of about 6 minutes per error. Correcting fuzzy lexical items indeed often requires domain documentation and expertise.


In our experimentation, the following questions, crucial to controlled natural language systems, have been considered: 
\begin{itemize}
\item  What are the strategies deployed by technical writers when they see the alerts? what do they think of the relevance of each alert?
how do they feel
about making a correction? How much they interact with each other ?
\item Over large documents, how is it possible to produce stable and homogeneous corrections? 
\item How much of the sentence is modified, besides the fuzzy lexical item? Does the modification
affect the sentence content?
\item How difficult is a modification and what resources does this requires (e.g. external documentation, asking someone else)?
How many corrections have effectively been done? How many are left pending and why?
\end{itemize}

\section{A Method for the definition of an Error Correction Memory}

Our analysis is based on a corpus of technical texts coming from seven companies, kept anonymous at their request. 
Our corpus contains about 120 pages from 27 documents. The main features considered to validate our corpus are:
(1) texts corresponding to various professional activities: product design, maintenance, production launch, specifications, regulations and requirements,
(2) texts following various kinds of business style and format guidelines imposed by companies,
(3) texts coming from various industrial areas: finance, telecommunications, transportation, energy, computer science, and chemistry.


The main principle is to observe technical writers when they make corrections from LELIE's alerts and to memorize any 
error with its final correction together with its precise context of utterance.
The absence of a correction is also memorized. After a certain period of observation, there is sufficient material to
develop the error correction memory. In addition, correctly realized utterances in the same context (i.e. without any alert) are also considered as a correction guide.

The main features and advantages of an error correction memory in the context of LELIE are:
\begin{itemize}
\item Corrections take into account the utterance context and the company's authoring practices, 
\item Corrections which are proposed after observation result from a consensus among technical writers in a group since an administrator (possibly via mediation)  determines the
best corrections to be kept given a context. These corrections are then proposed in future correction tasks in similar situations.
\item Corrections are directly accessible to technical writers: as a result, a lot of time is saved; furthermore, corrections become homogeneous over the various documents of the company,
\item Corrections reflect a certain know-how of the authoring habits and guidelines of a company, therefore they can be used to train novices.
\end{itemize}

This introduces a more dynamic and flexible view of implementing controlled natural language principles as suggested e.g. in (Ganier et al. 2007) than in standard authoring guidelines or boilerplates.

\subsection{A Lexicon of Fuzzy Lexical Items}

In the Lelie system, a lexicon has been implemented that contains the most common  fuzzy lexical items found in our corpus (about 450 terms). 
Since some items are a priori more fuzzy than others, a mark, between 1 and 3 (3 being the worse case) has been assigned
a priori. This mark is however not fixed, it may evolve depending on technical writers' behavior.
For illustrative purposes, Table 2 below gives figures about some types of entries of our lexicon for English. 
{\small 
\begin{center}
\begin{tabular}{|l|l|l|l|}
\hline 
category & number of entries & a priori severity level  \\  \hline
manner adverbs & 130 & 2 to 3 \\ \hline
temporal and location adverbs &  107 & in general 2 \\ \hline
determiners &  24 & 3  \\ \hline
prepositions & 31 & 2 to 3 \\ \hline
verbs and modals & 73 & 1 to 2 \\ \hline
adjectives & 87 & in general 1 \\  \hline
\hline
\end{tabular}
\end{center}
 Table 2.  Main fuzzy lexical classes.
}

\subsection{Memorizing Technical Writers' behavior}

An observation on how technical writers proceed was then carried out. 
The tests we made do not include any temporal or planning consideration (how much time it takes to make a correction, or how they organize
the corrections) or any consideration concerning the means and the strategies used
by technical writers. At this stage, we simply examine the correction results, which are stored in a database. At the moment, since no specific interface
has been designed, the initial and the corrected texts are compared once all the corrections have been made. The protocol to memorize corrections is the following:
\begin{itemize}
\item for a new fuzzy lexical item that originates an alert, create a new entry in the database, include its category and a priori severity level,
\item for each alert concerning this item, include it in its database entry with its context (see below) and the correction
made by the technical writer. Indicate
who made the correction (several writers often work on similar texts). Tag the term on which
the alert is in the input text and tag the text portion that has been changed in the resulting sentence. 
\item If the initial sentence has not been corrected then it is memorized and no correction is entered.
\end{itemize}

The database is realized in Prolog as follows:
{\small \begin{verbatim}
fuzzyitem([term], [category], [severity],
[[text fragment with alert, text after correction with tags, 
ID of writer], ....] ).
\end{verbatim}}
For example: {\small \begin{verbatim}
fuzzyitem([progressively], [adverb], [3],
[[[<fuzzy>, progressively, </fuzzy>, heat, the, probe], 
   [[heat, the, probe, <revised>, progressively, 
             in, 5, seconds, </revised>]],  [John] ] .... ] 
\end{verbatim} }

\subsection{Error Correction Memory Scenarios}

Considering technical writers corrections, error correction memory scenarios include the following main situations, which have been developed a priori and intuitively before
evaluating their operational adequacy:
\begin{enumerate}
\item A fuzzy lexical item that is {\bf not corrected} over several similar cases, within a certain word context or in general, no longer originates an alert. 
We are evaluating at the moment a threshold (e.g. 5 not corrected alerts) before this decision can be validated by technical writers.
The corresponding fuzzy lexical item in the
LELIE lexicon becomes inactive for that context, e.g. in {\it to minimize fire alarms}, 'minimize' describes a general behavior, not something very specific, 
it is therefore no longer displayed as an error.
Same situation for 'easy' in {\it a location that allows easy viewing during inspection}.
\item (2a) A fuzzy lexical item that is {\bf replaced or complemented by a value, a set of values or an interval}, may originate, via generalizations, 
the development of correction patterns that 
require e.g. values or intervals. 
For example, from examples such as:\\
{\it progressively close the pipe  $\rightarrow$ progressively close the pipe  in 30 seconds.\\
Progressively heat the probe $\rightarrow$ heat the probe progressively over a 2 to 4 mns period.\\
The power must be reduced progressively to 65\% to reach 180 knots $\rightarrow$ reduce the power to 65\% with a reduction of 10\% every 30 seconds to reach 180 knots}.\\
A correction pattern could be the association of progressively (to keep the manner and its continuous character) with a time interval, possibly complex,
as in the third example:\\
{\it progressively $\rightarrow$ progressively [temporal indication, type: value, interval, ...]}.\\
This pattern is composed of two facets: a relatively generic correction pattern that suggests a revised formulation (e.g. the adverb followed by an interval of values) and a {\bf correction recommendation} that
proposes, in context and when relevant, one or more typical precise values for the subfield 'value'. The pairing of these two levels generic / instantiation seems to be a good compromise between
adequacy and efficiency of the correction. 

(2b) In parallel with generalizing over corrections, the above item can be complemented by the {\bf observation of correctly realized utterance with the same context} (but no fuzzy term: e.g. {\it heat the probe in 2 to 4 mns}) 
via a direct
search in related texts. The idea is that errors are not systematic and that it may be possible to find correct realizations that may be used consistently with
the corrections that have been observed.
\item A fuzzy lexical item that is simply {\bf erased} in a certain context (probably because it is judged to be useless, of little relevance or redundant) originates a correction recommendation 
that specifies that it may not be necessary in that context.
For example: {\it  any new special conditions $\rightarrow$ any new conditions; proc. 690 used as a basic reference applicable to airborne $\rightarrow$ proc. 690 used
as a reference...}. In these examples, 'special' and 'basic' are fuzzy, but they have been judged not to convey a very heavy meaning, therefore they can be erased. 
\item A fuzzy lexical item may be {\bf replaced by another term or expression in context that is not fuzzy}, e.g. {\it aircraft used in normal operation $\rightarrow$ aircraft 
used with side winds below 35 kts and outside air temperature below 50 Celsius}, in that case the suggestion to revise 'normal' in context
is memorized and then proposed in similar situations. 
\item Finally a fuzzy lexical item may involve a complete rewriting of the sentence in which it occurs. This is the worst case, it should be avoided whenever possible because it often
involves changes in the utterance meaning.
\end{enumerate}

In a given domain, errors are very reccurent, they concern a small number of fuzzy terms, but with a large diversity of contexts.
A rough frequency indication for each of these cases, based on 52 different fuzzy lexical items with 332 observed situations 
can be summarized as follows:
{\small 
\begin{center}
\begin{tabular}{|l|l|l|}
\hline 
case nb. & number of cases & rate (\%)  \\  \hline
1 & 60 &  18 \\ \hline
2 &  154 & 46 \\ \hline
3 &  44 & 13  \\ \hline
4 & 46 & 14 \\ \hline
5 & 28 & 9 \\ \hline
\hline
\end{tabular}
\end{center}
Table 3. Correction situations distribution 
}

\subsection{Error Contexts}
Let us now define the parameters of these scenarios, namely: (1) definition of  contexts and (2) definition of generic patterns and specific correction recommendations.
In our first experiment, {\bf Contexts} are words which appear either before or after the fuzzy lexical item that characterize its context of utterance. In the case of fuzzy lexical items, a context is composed of nouns or noun compounds
(frequent in technical texts) $N_i$, adjectives $A_k$ and
action verbs $V_j$. Our strategy is to first explore the simple case of a fixed number of terms to unambiguously characterize a context, independently of the fuzzy lexical item category and usage. 
In our expriment, the context is composed of (1) a main or head word which is the word to which the fuzzy item applies
(e.g. 'fire alarms' in 'minimize fire alarms') and (2) additional words that appear either before or after the main one. 
The closest words in terms of word distance are considered.

In the context definition,
morphological variants are included and close words (sisters) if an ontology exists. The approach has the advantage of not including any syntactic consideration. 
To evaluate the number of additional words which are needed in the context besides the head word, we 
constructed 42 contexts from 3 different texts composed of 2, 3, 4 and 5 additional words. 
We then asked technical writers to indicate from what number of additional words each context was stable, i.e. adding a new words does not change what it means or refers to. 
Over our small sample, results are the following: 
{\small 
\begin{center}
\begin{tabular}{|l|l|}
\hline 
number of additional words & stability from previous set  \\  \hline
3 & 85\% \\ \hline
4 & 92\% \\ \hline
5 & 94\% \\ \hline
\hline
\end{tabular}
\end{center}
Table 4.  Size of context 
}

From these observations, a context of 4 additional words (if these can be found in the utterance) and the main words is adopted. 

\subsection{Error Correction Patterns}

Automatically defining {\bf correction patterns} from the different sets of samples in the database via generalization(s) would be the most straightforward approach.
However, we first want to evaluate the form and contents of patterns and recommendations that would be the most appropriate for efficiently correcting errors. 
The cooperation
between correction patterns and correction recommendations needs to be investigated. By efficiently correcting errors, we mean
adequacy w.r.t. (1) the error analysis and type of alert and (2) correction feasibility for the technical writer. 

For this first experiment, correction patterns have been defined manually considering
(1) the syntactic category of the fuzzy item and (2) the correction samples collected in the database.
A pattern is viewed as a guide which requires the expertise of the technical writer. It is not imperative.
Here are a few relevant and illustrative types of patterns, given in a readable form:
\begin{itemize}
\item  {\bf fuzzy determiners:} specification of an upper or a lower boundary (N) or an interval, e.g. pattern: {\tt [a few X]} $\rightarrow$ {\tt [less than N X]}, 
{\tt [most X]} $\rightarrow$ {\tt [more than N X]}. Besides patterns, which are generic, the context may induce a correction recommendation for the value
of X: depending on X and its usage (context) a value for X can be suggested, e.g. '12' in {\it take-off a few knots above V1 $\rightarrow$ take-off less than 12 knots above V1},
with Context = main term: knots, additional: take-off, above V1.
\item  {\bf temporal adverbs}, combined with an action verb, such as {\it frequently, regularly}: specification of a temporal value with an adequate quantifier, e.g.:  
{\tt [regularly Action]} $\rightarrow$ {\tt [every Time Action]}, where Time is a variable that is instantiated on the basis of the context or the Action.
An adverb such as {\it progressively} is associated with a Time interval when it modifies a durative verb:  
{\tt [progressively verb(durative)]} $\rightarrow$ {\tt [progressively verb(durative) in Time]}, e.g. {\it progressively close the pipe in 10 seconds}. Time is suggested by the correction recommendation level.
\item {\bf manner adverbs}, such as {\it carefully} which do not have any direct measurable interpretation, the recommendation is (1) to produce
a warning that describes the reasons of the care if there is a risk, or (2) to explain how to make the action in more detail, via a kind of 'zoom in', or (3) to simply skip the
adverb in case it is not crucial. For example,  {\tt [carefully Action]} $\rightarrow$ {\tt [carefully Action Warning]}, e.g. {\it carefully plug-in the mother card
$\rightarrow$ carefully plug-in the mother card otherwise 
you may damage the connectors}.
\item {\bf  prepositions} such as {\it near, next to, around, about} require the specification of a value or an interval of values that depends on the context. A pattern is for example:
 {\tt [near noun(location)]} $\rightarrow$ {\tt [less than Distance from noun(location)]}, where Distance depends on the context, e.g. {\it park near the gate $\rightarrow$ park less than 100 meters
 from the gate}. The variable Distance is contextual and constitutes the correction recommendation, making the pattern more precise.
\item {\bf adjectives} such as {\it acceptable, convenient, specific} as in {\it a specific procedure, a convenient programming language} can only be corrected via
a short paraphrase of what the fuzzy adjective means. For example, {\it convenient} may be paraphrased by {\it that has debugging tools}. Such paraphrases can
be suggested to technical writers from the corrections already observed and stored in the database that implements the error correction memory.
\end{itemize}

At the moment, 27 non-overlapping patterns have been defined to correct fuzzy lexical items. 
Some patterns refer to frequent errors, with stable corrections: they can be induced and validated after about 80 pages of corrected text. Others are less frequent and 
require much larger text volumes.
Error correction recommendations are more difficult to stabilize because contexts may be very diverse. At the moment, (1) either a precise recommendation has emerged or has been found in correct texts and has 
been validated and is proposed or (2)
the system simply keeps track of all the corrections made and displays
them by decreasing frequency. 

We are now defining a protocol to evaluate the adequacy of these
patterns w.r.t. the document contents and their usability by technical writers. Adequacy is related to the linguistic and contents level: the principle is that the
meaning of the corrected utterances must not be affected or in a very minimal way by the changes suggested by correction patterns. Usability means that the patterns
and the correction recommendations that make them more precise can be understood and used by technical writers after a short training period and that they
really use them in the long range, over several types of documents.

\section{Perspectives}

In this paper, we have explored the notion of error correction memory, which, paired with the LELIE system that detects specific errors of technical writing, allows
both the detection and the correction of errors. Correction scenarios are based on an architecture that develops an error correction
memory based on (1)  generic correction patterns and (2) correction recommendations for elements in those patterns which are more contextual. Both levels are acquired from
the observation of already realized corrections and correct texts.
This approach is quite new, it needs an in-depth evaluation in terms of linguistic adequacy and usability for technical writers.

In parallel with fuzzy items, we are exploring other facets of an error correction memory for other major types of errors such as negation or complex sentences.
For complex sentences there are situations which can be handled quite straightforwardly because they reflect a stable writing practice. For example, illustrations, exceptions,
purposes or circumstances
can be realized in one or more additional sentences instead of being inserted into the main one. For example, roughly, a pattern for purpose clause 'externalization' can be for a requirement:
{\tt [X shall Y conj(purpose) Z]} $\rightarrow$ {\tt [X shall Y. The goal is to Z]}. We believe that a similar technique could be used for heavy sequences of N+N or noun complements and
heavy sequences of coordination or relatives. 

\end{document}